# A general-purpose AI assistant embedded in an open-source radiology information system

Saptarshi Purkayastha[1], Rohan Isaac[1], Sharon Anthony[1], Shikhar Shukla[1], Elizabeth A. Krupinski[3], Joshua A. Danish[2], and Judy W. Gichoya[3]

[1] Indiana University-Purdue University Indianapolis, Indianapolis IN 46202, USA
{saptpurk, risaac, shaantho, shikshuk}@iu.edu
[2] Indiana University Bloomington, Bloomington IN 47405, USA
jdanish@indiana.edu
[3] Emory University, Atlanta GA 30322, USA
{ekrupin, judywawira}@emory.edu

**Abstract.** Most AI models developed in research settings or Kaggle competitions do not translate to clinical application because they are difficult to validate or embed in real-world clinical practice. Radiology AI models have made significant progress in near-human performance or surpassing it. However, AI model's partnership with human radiologist remains an unexplored challenge due to the lack of health information standards, contextual and workflow differences, and data labeling variations. To overcome these challenges, we integrated an AI model service which uses DICOM standard SR annotations into the OHIF viewer in the open-source LibreHealth Radiology Information Systems (RIS). In this paper, we describe the novel Human-AI partnership capabilities of the platform, including few-shot learning and swarm learning approaches to retrain the AI models continuously. Building on the concept of machine teaching, we developed an active learning strategy within the RIS, so that the human radiologist can enable/disable AI annotations as well as "fix"/relabel the AI annotations. These annotations are then used to retrain the models. This helps establish a partnership between the radiologist user and a user-specific AI model. The weights of these user-specific models are then finally shared between multiple models in a swarm learning approach. We discuss the potential advantages and pitfalls for such assemblage learning through the integration of AI models into clinical workflows.

## 1 Introduction

Integration of artificial intelligence (AI) into radiology has been the topic of debates in the last several years - ranging from suggestions to "not train any more radiologists since their work is redundant" to AI demonstrating superhuman performance for various diagnostic tasks [10,17]. As progress has been made to understand how AI can be used in radiology, some experts predict that radiologists who use AI will replace those who do not use AI [9]. Despite differing opinions about how the work of radiologists will evolve in an AI world, it is clear that the radiologists' professional activities will change due to these changes [23].

The digital era in medicine was pioneered by radiology, which was among the first medical specialties to adopt computer-based technologies through implementation of the picture archiving and communication system (PACS) workflows for image interpretation. While the introduction of new technologies was initially seen as a means of producing better images, technology has fundamentally altered how images are acquired, displayed stored, analyzed and shared, resulting in a significant change in some of the tasks radiologists engage in clinically [10]. The radiologist's job encompasses various aspects, including image interpretation and analysis, report creation, and patient and doctor consultation [7]. The diagnostic process typically includes detection and characterization of findings in medical images. [10].Radiologists are considered among the most technologically savvy healthcare practitioners [10].

The process of generating these radiology images requires process management - from the time a study is ordered, to when its performed, and a report dictated [15]. Due to increased utilization of imaging in patient triage across multiple departments, there has been increasing demand for better productivity and efficient workflow management [13]. The radiology department manages a complicated workflow involving various personnel, different technologies, and time-sensitive information to offer clinical services to referring doctors and patients [7].

In this context, AI in radiology holds promise to not only improve the diagnostic/interpretive tasks but to improve workflow, increase efficiency and enhance quality of service [19]. The realization of this impact of AI remains elusive because the real-world implementation of many AI models is fraught with multiple challenges [12], such as dataset variability [11], lack of adequate testing [20], variation in practices [5] and limited tooling that integrates AI with existing workflows [16,3]. Many have argued that better partnership between human radiologists and AI models is a way to remove these implementation challenges and build end user trust in AI [4,24]; but few in-situ implementations, if any, can be found in literature [25].

In this paper, we describe updates to LibreHealth Radiology (LH-radiology), a well-regarded standards compliant [6,14], open-source Radiology Information System (RIS), which integrated an AI model service into the full workflow of an RIS in implementation. The LibreHealth RIS has multiple features that manage the radiology workflow including order entry and HL7 order import, acquisition and modality management, connectivity to multiple PACS systems such as Orthanc [8] and DCM4CHEE; and an integrated open-source DICOM Viewer called Open health imaging foundation (OHIF) [26]. The LibreHealth AI model service is designed with Human-AI partnership in mind, where a human radiologist can retrain a model and live updates to the AI model can be shared with other radiologists on the network. The AI Model Service enables two-way communication supporting training and inference between different radiology AI models in the LibreHealth RIS. This allows for integration of AI enabled functionality including disease classification, body organ segmentation, automated study lists, volumetric measurement, image enhancements and decision support tools into the regular radiologist workflow. The embedded OHIF viewer displays the model outputs and can be used for advanced functionalities such as prioritizing patient study lists, assigning them to authorized users, and scheduling different modalities for capturing.

## 2   LibreHealth RIS Architecture

The AI-integrated LibreHealth RIS architecture consists of 2 main components - RIS module (OMOD) and AI model service. This system is built on an Electronic Health Records (EHR) toolkit that provides the foundation for API and data model that is utilized by many health IT systems. The LH-radiology, AI model service, and supporting components are shown in Figure 1.

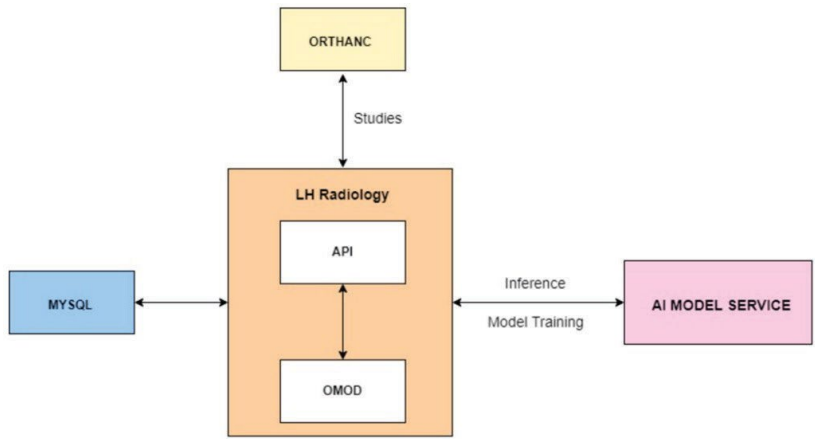

**Fig. 1.** The LibreHealth RIS Architecture

Once an image has been acquired, the study list is imported from the Orthanc PACS and displayed in the OHIF study list in the LH-radiology module. The information is stored in the MYSQL database which serves as the repository. When the radiologist accesses the study list to work on a study, the AI model service is called.

### 2.1   LibreHealth Radiology Module:

The LH-radiology module was created using a user-centered participatory design with multidisciplinary teams in human-computer interaction, computer scientists, radiologists and general physicians who collaborated on user stories to create features of the module.The LH-radiology system adds the abilities of a Radiology Information System (RIS) into the LibreHealth Toolkit. This adds functionality to the EHR to manage radiology studies, review orders, configure web configuration for PACS and viewer connectivity, support REST API-based communication with the AI model service and the MySQL database. Standards-compliant interfaces with HL7, DICOM and WADO are some of the main strengths of the RIS.

## 2.2   LibreHealth Radiology Artificial Intelligence Model Service:

The AI model service is a REST API-based service that is used to communicate with AI models from LibreHealth RIS. Through this method, a two-way communication i.e., training and inference is enabled between different AI models deployed in the LibreHealth RIS. The RIS-embedded DICOM Viewer called OHIF presents the outputs of the models. The AI model service is able to run as a standalone system that can receive DICOM Modality Worklist (MWL) or DICOM images and return AI model output as a JSON file. These JSON results are then used to prioritize patient study lists and assign them to different radiologists.

The model service also offers a comprehensive set of endpoints that can be used to retrieve a list of all the models available and their inference results including multi-label and multi-class classification with individual class probabilities, bounding boxes around regions of interest for each label (localization), and border highlighting of regions of interest (semantic segmentation). The AI Model Service also supports the retraining of the models when their outputs are modified by a radiologist.

## 3   METHODS

### 3.1   OHIF viewer and its extensions

The OHIF medical imaging viewer is embedded in the LH-radiology module. OHIF provides access to many functions including ability to retrieve and load images from most PACS sources and formats; render sets in 2D, 3D, and reconstructed representations; support the manipulation, annotation, and serialization of observations; supports internationalization, OpenID Connect, offline use, hotkeys, and many more features [26].

The OHIF viewer in LH-radiology is configured with a custom extension that allows 2-way communication with the AI model service using DICOM-SR annotations. DICOM-SR provides a standard way to annotate findings by radiologists when they interpret studies. These annotations are stored in their own layer that is overlayed over the original image, without any modifications to the original image. Our custom extension along with core extensions from OHIF that are useful for measurement tasks are installed. These core extensions include:

1. @ohif/extension-cornerstone (The core OHIF Viewer's toolbar)
2. @ohif/extension-dicom-html (To render DICOM images on a canvas)
3. @ohif/extension-dicom-microscopy (For image manipulation functionality)
4. @ohif/extension-dicom-pdf (To save and print as PDF from DICOM)
5. @ohif/extension-vtk (Image processing, 3D graphics, and volume rendering)
6. @ohif/extension-LH-radiology (2-way communication with AI model service)

## 3.2   AI Model Service

The AI model service is a general-purpose AI assistant that works with a variety of AI models. The AI model service uses REST API endpoints to communicate with the OHIF viewer. When a radiologist interacts with the images from the study list, a GET pull is made from the model service to load prior DICOM-SR annotations, along with a GET call to the PACS system to retrieve the radiology image along with user-saved DICOM-SR annotations that are independent from the AI generated annotations. The AI model service then performs the requested operations such as inference, training, model updating and returns the model output to the OHIF viewer. The OHIF viewer receives this output and displays the model inference, model version and model status (retraining/swarm learned) on the user interface. A sample GET request format is shown in Table 1 made from the viewer to obtain information of the AI models in the model service along with their status.

**Table 1.** GET bounding-box API endpoint

| Request | Response |
|---|---|
| GET /bounding-box | { <br> "data": [{ <br>     "model": "string", <br>     "version": "string" <br> }], <br> "status": "string" <br> } |

The response provides information on the AI models available for inference for different modalities such as chest x-ray, mammography and others. A GET request payload (containing the image Base64 encoded string and model information) made from the viewer returns back the AI model inference result which contains bounding box coordinates, image annotation, the model applied for inference and its status.

If the radiologist feels that the AI model inference is inaccurate, the co-ordinates of the bounding boxes on the OHIF viewer can be adjusted generating new annotations that are used to retrain the model. In this case, the AI model service receives a POST request as shown in Table 2 containing the new bounding box coordinates as payload, along with the model version to retrain. The retraining is planned as a batch procedure so that the AI model is retrained using annotations from multiple users simultaneously. During the training process, the images used for inference on which we have new annotations are used for retraining, and the weights of the pretrained AI model are updated. After the training process is complete, a new version of the AI model is created and is ready for inference. We deployed the CheXnet [17] model to demonstrate the LibreHealth radiology AI enabled workflow as shown in Figure 2. On the chest Xray, the original image is rendered in the OHIF viewer with corresponding radiologist and AI annotations shown as bounding boxes.

**Table 2.** POST model-update API endpoint

| Request | Parameters | Response |
|---|---|---|
| POST /model-update | {<br>  "annotationText":"string",<br>  "image": "string",<br>  "model": "string",<br>  "modelVersion": 0,<br>  "x1": 0,<br>  "x2": 0,<br>  "x3": 0,<br>  "x4": 0,<br>  "y1": 0,<br>  "y2": 0,<br>  "y3": 0,<br>  "y4": 0<br>} | {<br>  "data": [{<br>    "model": "string",<br>    "modelVersion": "string"<br>  }],<br>  "status": "string"<br>} |

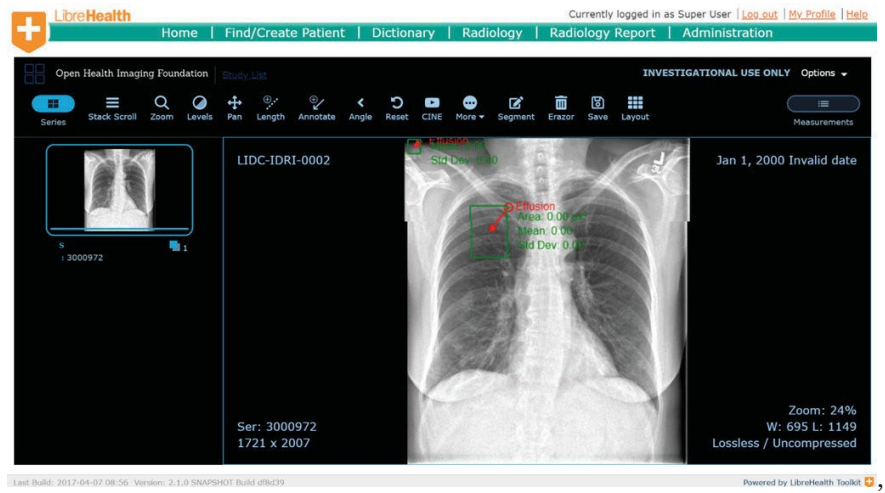

**Fig. 2.** LH-radiology ChexNet-produced annotations in OHIF Viewer

The design of the AI model service is modular, allowing for future extensions in its functionality. For example, we have been able to use this functionality to deploy the AI results to prioritize studies on the worklist.

### 3.3   Swarm Learning and Few Shot Learning

While the AI model service allows the ability to customize the model to the preference of a specific radiologist determined by a user's annotations and corrections, this personalization may result in model bias. The personalized AI model may aggregate errors that the human radiologist makes regularly or might contain errors that the human radiologist regularly fails to correct in the model. To reduce model bias while maintaining the model personalization, we share knowledge across multiple human-AI partnership-based models by sharing model weights through the AI model service. Swarm learning is utilised in the AI Model Service where models are built independently on private data for each individual radiologist [22] as shown in Figure 3. The trained model weights are then shared with other personalized radiologist models via a swarm application interface and merged to create an updated model with updated parameters [18]. The resulting model has accurate inferences equivalent to or better compared to a model trained on a central site on all gathered data.

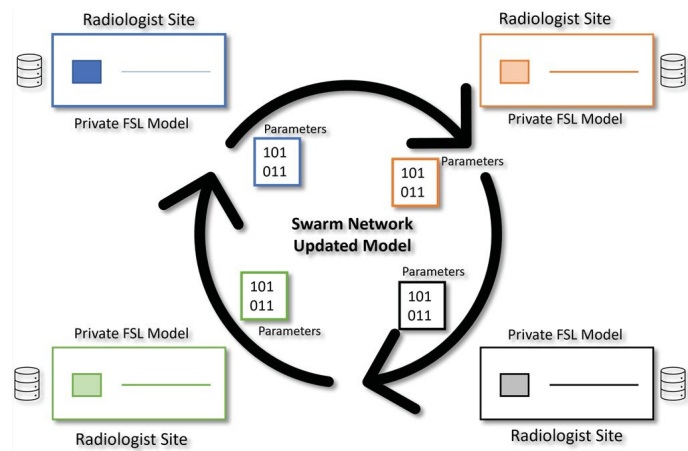

**Fig. 3.** LH-radiology Swarm Network Architecture

Swarm learning in the AI model service enables collaboration among radiologists while maintaining data confidentiality. The specific layers of the updated model are manipulated using the collected model parameters using an ensemble technique [1] where the model parameters of the final model are estimated through an additive or weighted method. The AI model service performs model personalization with each correction that the human radiologist makes to the model output. As discussed earlier, these are batched and processed. However, since there are very few images available for retraining, fine tuning the model often results in bias and model performance degradation. To counter this, we use few-shot learning, a process of using limited newly labeled images or new labels on existing images, to adjust model weights to these new training set. In our case, a small sample of erroneous inference images is gathered after a model has been trained and validated, and the model is then retrained using image triplets (a false positive or false negative, a true positive, and a true negative) [2]. With a few epochs and a small number of images, the retrained few-shot learning model achieves noticeable performance improvements [2]. Additionally, few-shot learning creates opportunities for rapid model retraining in human-in-the-loop systems, allowing a radiologist to swiftly relabel incorrect judgments [21].

## 4    DISCUSSION

In this work, we successfully demonstrate an end to end deployment of an AI inference and training service implemented in the usual workflow of professional radiologists. This work is novel because of the following contributions: 1) the design and implementation of the system to support multiple modalities and multiple models for training and inference; 2) demonstration of personalization in AI development and inference to support active learning through the swarm application interface; and 3) a standards based approach to integrate multiple systems. This makes the platform support multiple use cases necessary to understand how radiologists can work with AI. For example, detection and segmentation tasks on radiology images are matched to how radiologists view images and interpret them. Therefore, a classification task e.g, 80% chance of pneumothorax without localization can be difficult to explain. To validate such a use case on LH-radiology would be straightforward if a classification and detection model are available and inference performed on one image. The real time feedback for the model performance lowers the burden of annotation in batches, and can produce better quality annotations compared to weakly supervised labeling techniques that rely on natural language processing of radiology reports.

Another innovation of our system design is the opportunity to evaluate the personalized AI models trained on specific radiologists annotations. These swarm evaluations can be evaluated for big trends and radiologist preferences when interacting with model outputs - providing an opportunity to understand desirable configurations that may be required during AI deployments to minimize the disruption of the radiologist. For example, if the radiologists do not interact with model outputs for normal exams, a configuration could involve turning off the model output display for normal cases while the model service is running in the background. Similarly, a blind spot observed across several radiologists but supplemented by AI use can be used to train junior radiologists or provide refresher classes for more experienced radiologists.

## 5    CONCLUSION

In this paper, we discussed the process and significance of the AI-driven Libre-Health RIS using open source components. Incorporating different perspectives in the system design, we developed an architecture that allows for AI deployment in the usual radiologist workflow working with multiple model types and image modality types by taking advantage of existing imaging standards. Our system provides a real-world replication of how to integrate AI into the radiologist workflow using a personalized approach. More accurate and reproducible radiology assessments can potentially be made with the help of this AI integration, but this needs to be verified with human observer studies. We welcome readers to become part of our community to test and participate in improving the system to support better infrastructure to validate AI models to multiple end-users.

**Acknowledgements** This work was funded by the US National Science Foundation (#1928481) from the Division of Electrical, Communication & Cyber Systems.